\documentclass{article}
\usepackage{spconf,amsmath,epsfig}

\usepackage[pagebackref=true,breaklinks=true,letterpaper=true,colorlinks,bookmarks=false]{hyperref}
\usepackage{times}
\usepackage{epsfig}
\usepackage{graphicx}
\usepackage{amsmath}
\usepackage{amssymb}
\usepackage{multirow}
\usepackage{multicol}
\usepackage{bm}
\usepackage{xcolor}
\usepackage{adjustbox}
\usepackage{booktabs}

\title{ChangeBind: A Hybrid Change Encoder for Remote Sensing Change Detection}
\name{Mubashir~Noman$^1$  \quad
        Mustansar~Fiaz$^2$  \quad
        Hisham~Cholakkal$^1$
        \vspace{-0.5em}
        }
\address{$^1$Mohamed bin Zayed University of AI \quad $^2$IBM Research
}

\begin{document}
%
\maketitle
\begin{abstract}
Change detection (CD) is a fundamental task in remote sensing (RS) which aims to detect the semantic changes between the same geographical regions at different time stamps.  Existing convolutional neural networks (CNNs) based approaches often struggle to capture long-range dependencies. Whereas recent transformer-based methods are prone to the dominant global representation and may limit their capabilities to capture the subtle change regions due to the complexity of the objects in the scene. To address these limitations, we propose an effective Siamese-based framework to encode the semantic changes occurring in the bi-temporal RS images.  The main focus of our design is to introduce a change encoder that leverages local and global feature representations to capture both subtle and large change feature information from multi-scale features to precisely estimate the change regions. Our experimental study on two challenging CD datasets reveals the merits of our approach and obtains state-of-the-art performance. Code is available at \url{https://github.com/techmn/changebind}.
\end{abstract}
\section{Introduction}
\label{sec:intro}

The accelerated growth in urbanization necessitates the urge of land management to avoid the adverse environmental and socioeconomic effects \cite{Nuissl2021_urbanisation_and_land_use, pang2022cd}. Recent advancements in remote sensing (RS) technology have facilitated researchers to utilize deep learning techniques, such as change detection, for effective management of land use. 
Change detection (CD) refers to the problem of identifying the relevant semantic changes between bi-temporal remote sensing images ~\cite{shi2020change, elgcnet2024, kucharczyk2021remote, noman2023remote}.
Here, the relevant semantic changes represent the construction-related changes such as building construction or demolishing and other man-made facilities like road constructions.
Hence, CD plays a vital role in various remote sensing applications such as land usage monitoring \cite{fonseca2021pattern,wen2021change}, land resource management \cite{yin2021integrating, chughtai2021review}, disaster assessment \cite{zheng2021building}, forestry \cite{slingsby2020near}, and more.

Earlier on, several works utilized the conventional machine learning approaches for the CD task including support vector machines \cite{migas2021assessment}, random forest \cite{zhu2014continuous, ghosh2011fuzzy}, principal component analysis (PCA) \cite{hussain2013change}, change vector analysis (CVA) \cite{tewkesbury2015critical}, and other methods. However, these methods mainly rely on hand-crafted features and struggle to detect change regions in RS images due to the challenges posed by seasonal variations, brightness and illumination variations, appearance variations, and the presence of irrelevant objects. In addition, the detection of change regions having varying shapes and sizes is another challenging task. 

\begin{figure*}[t!]
\centering
 \includegraphics[width=0.95\textwidth]{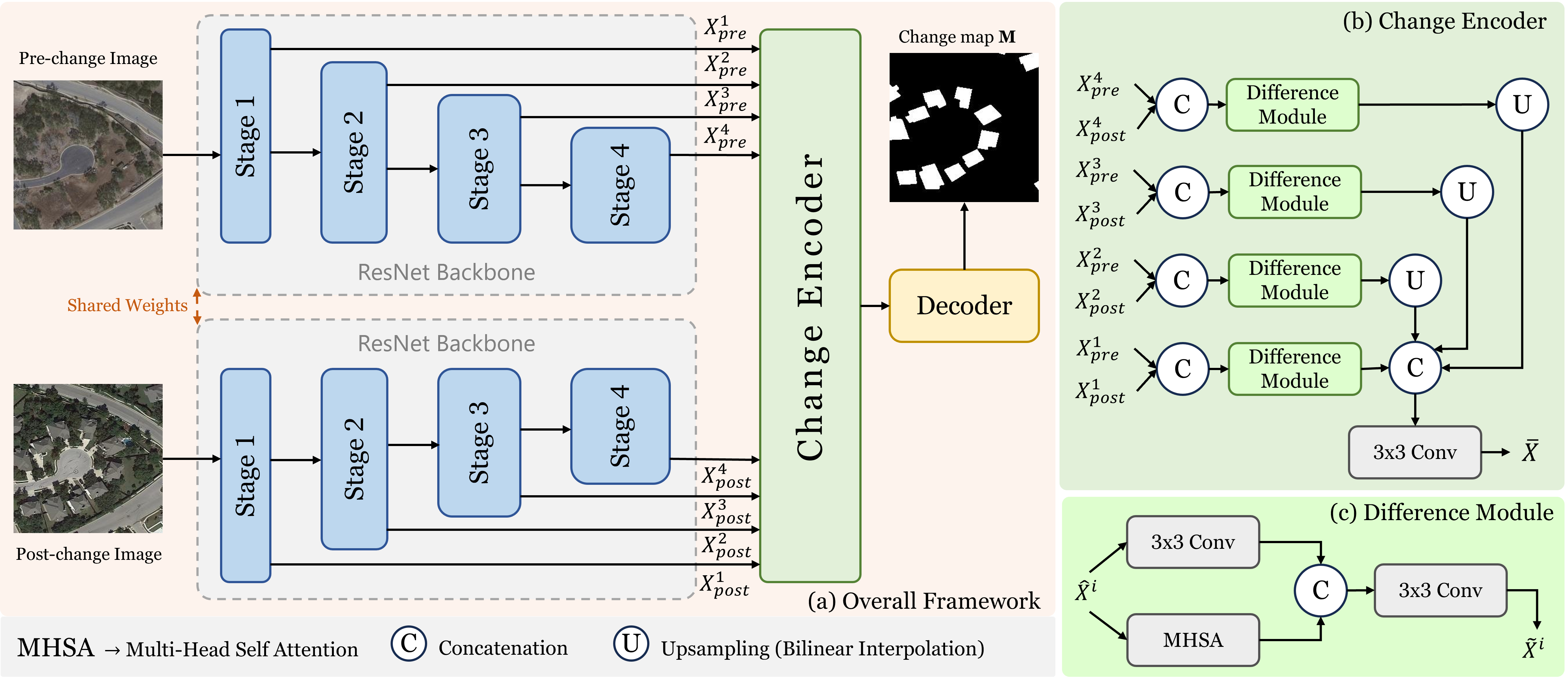} 
\caption{ The (a) 
 illustrates the overall architecture of our proposed CD framework, referred as ChangeBind. The model takes a pair of bi-temporal images and extracts multi-scale features through a Siamese-based ResNet backbone. The multi-scale features ($X_{pre}^i$ and $X_{post}^i$, where $i \in {1,2,3,4}$) are fed to the change encoder that highlights the semantic change regions. Afterward, a decoder is utilized to upsample the encoded change features and predict change map $M$. The (b) represents the structure of the change encoder that takes features $X_{pre}^i$ and $X_{post}^i$, and utilizes the difference module to encode change regions. The (c) shows the design of the difference module which takes concatenated features of a single scale level, and utilizes convolution to obtain convolutional change encodings (CCE) and MHSA for obtaining attentional change encodings (ACE). These CCE and ACE feature representations are merged and projected using a convolution operation within the difference module. The outputs of the difference modules at higher scale levels are upsampled and combined to obtain the encoded change ($\bar X$)  representations. Finally, these representations are input to the decoder to obtain the change prediction mask. }
 \label{fig:main_framework}
\end{figure*}

Recently, deep learning-based CD methods \cite{chen2020dasnet, fang2021snunet, chen2021remote_bit, yan2022fully_ftn, bandara2022transformer, li2022transunetcd, ke2022hybrid_transcd} have demonstrated admirable CD performance compared to the traditional CD methods by utilizing the convolutional neural networks (CNNs) and transformers. 
For instance, BIT \cite{chen2021remote_bit} utilizes ResNet \cite{kaiming2016_resnet} backbone to extract features from the bi-temporal images, concatenates the two feature representations, apply the self-attention to capture the global contextual relationships, and take feature difference to encode the changes between the bi-temporal images. Alternatively, ChangeFormer \cite{bandara2022transformer} uses self-attention in the Siamese-based encoder to enlarge the receptive field for extracting the representations of the objects having various sizes. Afterward, the extracted  features from the two images are concatenated and fed to the convolution layers to encode the change regions. Similar to BIT \cite{chen2021remote_bit}, Li et. al. \cite{li2022transunetcd} utilize self-attention to encode the change regions present in the bi-temporal features.
After extracting features from the Siamese-based backbone, existing methods either utilize a self-attention mechanism or convolution operation to encode the difference between the two feature representations. Although both convolution and self-attention operations are quite effective in capturing the changes present in the RS images. However, the dominance of global contextual representations in self-attention may restrict its ability to detect the subtle change regions. Alternatively, convolutions are impressive in capturing the textural and fine details but may not accurately capture the large changes due to the small receptive fields. Therefore, it is desirable to utilize both convolutions and self-attention operations for accurately detecting the subtle and large change regions. To this end, we propose a \textit{simple yet effective} method that efficaciously encodes the change regions of varying sizes present in the bi-temporal RS images. In summary, our contributions are:
\begin{itemize}
\item We propose an effective way of exploiting the feature representations of bi-temporal RS images for the change detection task.
\item The proposed approach efficaciously captures the subtle as well as large changes by utilizing the benefits of both convolutions and self-attention operations along with the multi-scale feature information.
\item We demonstrate the practicality of the our approach by experimenting on two challenging CD datasets. 
\end{itemize}

\begin{table*}[t!]
\centering
\caption{{State-of-the-art comparison on LEVIR-CD and CDD-CD datasets in terms of F1, IoU, and OA metrics. Our method demonstrates superiority compared to existing methods and obtains state-of-the-art performance. The best two results are in red and blue, respectively.
}}
\label{tbl:comaprison_on_LEVIR_CDD}
\setlength{\tabcolsep}{20.0pt}
\scalebox{1.0}{
\begin{tabular}{l|ccc|ccc} \hline
\multicolumn{1}{l|}{\multirow{2}{*}{Method}}   &\multicolumn{3}{c|}{LEVIR-CD}  & \multicolumn{3}{c}{CDD-CD}   \\  \cline{2-7} 
\multicolumn{1}{l|}{} &   F1   & OA & IoU  & F1   & OA & IoU  \\  \cline{1-1} \hline \hline 
FC-Siam-Diff   \cite{daudt2018fully} & 86.31 & 98.67 & 75.92 & 70.61 & 94.95 & 54.57 \\
DASNet    \cite{chen2020dasnet}  & 79.91 & 94.32 & 66.54 & 92.70 &  \textcolor{blue}{98.20} & 86.39 \\
DTCDSCN  \cite{liu2020building} & 87.67 & 98.77 & 78.05 & 92.09 & 98.16 & 85.34 \\

STANet   \cite{chen2020spatial}  & 87.30 & 98.66 & 77.40 & 84.12 & 96.13 & 72.22 \\
BIT  \cite{chen2021remote_bit}   & 89.31  & 98.92 & 80.68  & 88.90 & 97.47 & 80.01  \\ 
ChangeFormer  \cite{bandara2022transformer}  & 90.40 & \textcolor{blue}{99.04} & 82.48  & 89.83  & 97.68 &  81.53 \\
TransUNetCD  \cite{li2022transunetcd}  & \textcolor{blue}{91.11}  & -- & \textcolor{blue}{83.67}  & \textcolor{blue}{97.17}  & -- &  \textcolor{blue}{94.50}  \\
\hline
\textbf{Ours}  & \textcolor{red}{91.86} & \textcolor{red}{99.18} & \textcolor{red}{84.94}  &  \textcolor{red}{97.65} &  \textcolor{red}{99.44} &  \textcolor{red}{95.41} \\  \hline
\end{tabular}}
\end{table*}

\section{Method}
\label{sec:method}
In this section, we discuss the limitations of the baseline approach and the proposed method in detail.
\subsection{Baseline Framework}
\label{ssec: baseline}
We adapt the recent BIT \cite{chen2021remote_bit} method as our baseline framework since it has an unsophisticated three-phase architecture and provides promising results. In the first phase, the base framework utilizes a Siamese-based ResNet \cite{kaiming2016_resnet} backbone to extract features from the bi-temporal images. Next, the extracted bi-temporal features are concatenated and transformer blocks are used to enlarge the receptive fields and obtain rich feature representations. Finally, the enhanced features are split and the absolute difference is taken between these features followed by a prediction head to detect change regions.

\noindent \textbf{Limitations: } The base framework provides promising results, however, it does not fully exploit the multi-scale information extracted through the ResNet backbone. Additionally, the utilization of self-attention on the concatenated feature representations may emphasize more on the large regions compared to the small regions. As a result, the model strives to detect subtle change regions, and its performance is degraded. To alleviate the above limitations, we propose a hybrid change encoder that benefits from the properties of both convolutions and self-attention and provides better detection results for subtle and large change regions. Furthermore, motivated by FPN \cite{lin2017feature}, we utilize the multi-scale information extracted from the backbone network and separately encode the change information at four scale levels thereby improving the model performance.

\subsection{Overall Architecture}
\label{ssec: overall_arch}
The overall architecture of the proposed framework is illustrated in Fig. \ref{fig:main_framework}. The proposed framework takes a pair of bi-temporal RS images, referred as pre-change image $I_{pre}$ and post-change image $I_{post}$, and utilizes a Siamese ResNet backbone to extract multi-scale features at four scale levels i.e., $X_{pre}^i$ and $X_{post}^i$ where $i \in {1,2,3,4}$ as depicted in Fig. \ref{fig:main_framework}-(a). The extracted features $X_{pre}^i$ and $X_{post}^i$ are fed to the change encoder that separately processes the multi-scale features and finally combines them through a convolution layer to obtain rich feature encodings $\bar X$. Lastly, the encoded change representations are input to a decoder that upsamples the features by utilizing a transpose convolution layer followed by a residual convolutional block for feature enhancement. The decoder utilizes two transpose convolution layers to obtain the same spatial resolution as of the input image. Lastly, a convolution layer is used to obtain the change map $M$ prediction. Next, we discuss the change encoder in detail.

\subsubsection{Change Encoder}
\label{sssec: change_enc}
As illustrated in Fig. \ref{fig:main_framework}-(b), the multi-scale representations ($X_{pre}^i$  and $X_{post}^i$) of the bi-temporal images are fed to the change encoder. At each scale level, the feature representations are concatenated to obtain $\hat X^i$ and input to the difference module that aims to highlight the large as well as subtle change regions. Unlike the base framework, here we utilize the concatenation operation instead of the difference operation aiming the model to automatically learn the difference between the two representations. The difference module, as shown in Fig. \ref{fig:main_framework}-(c), utilizes a $3 \times 3$ convolution to highlight the fine details and obtain the convolutional change encodings (CCE). Meanwhile, it applies the multi-head self-attention (MHSA) operation to obtain global contextual representations or attentional change encodings (ACE) for focusing on the large change regions. Consequently, we fuse the two encodings extracted through convolution and MHSA operations through concatenation. Later, these fused features are realized with a $3 \times 3$ convolution layer to obtain change encodings at a single scale level. The encodings of the second, third, and fourth scale levels are upsampled through bi-linear interpolation to obtain a similar spatial resolution to the first scale for concatenation operation. Finally, the concatenated multi-scale encodings are merged by utilizing a $3 \times 3$ convolution to obtain rich change representations which are then passed to the decoder for change map prediction.

\begin{figure*}[!t]
\begin{center}
{\includegraphics[width=1\linewidth, keepaspectratio] {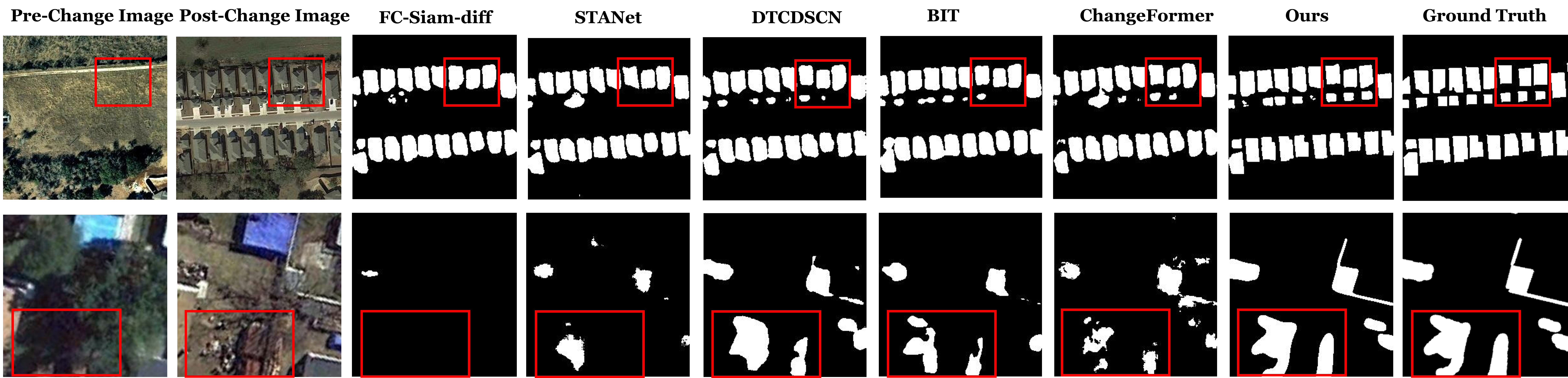}}
\end{center}
\caption{Qualitative results on the LEVIR-CD (top row) and CDD-CD (bottom row) datasets. We present a comparison with the best five existing change detection methods in the literature, whose codebases are publicly available. The highlighted region shows that our method is better at detecting the change regions as compared to  FC-Siam-diff \cite{daudt2018fully}, STANet \cite{chen2020spatial}, DTCDSCN \cite{liu2020building},  BIT \cite{chen2021remote_bit}, and ChangeFormer \cite{bandara2022transformer} methods.}
\label{fig_levir_cdd_vis}
\end{figure*}

\section{Experimental Section}
\label{sec:exp}

\subsection{Datasets and Evaluation Protocols:} In this work, we utilize two challenging CD datasets to verify the performance of our framework. 

\noindent \textbf{\textit{LEVIR-CD} \cite{chen2020spatial}:} is a publicly available CD dataset comprising of 637 high-resolution (0.5m per pixel) images of size $1024 \times 1024 \times 3$ collected from the Google Earth. Similar to the other works \cite{bandara2022transformer, chen2021remote_bit}, we utilize the cropped version of dataset having the spatial size of $256 \times 256 \times 3$ and default data splits of train, val, and test equal to 7120, 1024, and 2048, respectively.

\noindent  \textbf{\textit{CDD-CD } \cite{Lebedev2018CHANGEDI}:} is another public CD dataset having seasonal variations. 
 The dataset is available in non-overlapping cropped patches of size $256 \times 256 \times 3$. Similar to \cite{li2022transunetcd}, we utilize the cropped version and default data split of 10000, 3000, and 3000, for train, val, and test sets respectively. 
 
\noindent \textbf{\textit{Evaluation Protocols:}} Following other works \cite{bandara2022transformer, yan2022fully_ftn}, we evaluate the proposed method in terms of   \textit{change class} F1-score, \textit{change class} Intersection over Union (IoU) and overall accuracy (OA) metrics for both datasets.
 
\subsection{Implementation Details:}
The proposed method is implemented in PyTorch utilizing 4 A100 GPUs. Our network takes a pair of bi-temporal RS images of size $256 \times 256 \times 3$ and generates a binary change mask $\bm M$ which is calculated via pixel-wise argmax operation along the channel dimension. We use pre-trained ResNet50 \cite{kaiming2016_resnet} model as a backbone network to generate multi-stage features, which are input to our proposed changebind encoder. During training, the model is optimized using pixel-wise \textit{cross-entropy} loss and \textit{AdamW} optimizer is employed having a weight decay of 0.01 and beta values equal to (0.9, 0.999). We utilize a batch size of 16 for single GPU. We initialize the learning rate to 3e-4 and trained the model for 200 epochs. The value of the learning rate is decreased linearly till the last epoch.

\begin{table}[t!]
\centering
\caption{Ablation study on the LEVIR-CD dataset. 
The best two results are in red and blue, respectively.
}
\label{tbl:ablation_levir}
\scalebox{0.9}{
\begin{tabular}{l|ccc} \hline
Method & F1 & OA & IoU  \\
 \hline \hline
Baseline & 90.76 & 99.09 & 83.08 \\ 
Baseline + MSF  & 91.17 & 99.13 & 83.77 \\
Baseline + MSF + CCE & 91.51 & 99.15 & 84.34 \\  
Baseline + MSF + ACE  & \textcolor{blue}{91.62} & \textcolor{blue}{99.16} & \textcolor{blue}{84.54} \\
\hline
Baseline + MSF + CCE + ACE (Ours) & \textcolor{red}{91.86}  & \textcolor{red}{99.18}   & \textcolor{red}{84.94} \\ \hline
\multicolumn{4}{l}{
\footnotesize{$^*$MSF refers to multi-scale feature fusion.}}
\end{tabular}}
\end{table}

\subsection{Quantitative Comparison:}
In Table \ref{tbl:comaprison_on_LEVIR_CDD}, we present a quantitative comparison of our method with LEVIR-CD and CDD-CD datasets.
We compared our method with both CNN-based (FC-Siam-Diff \cite{daudt2018fully}, DASNet \cite{chen2020dasnet}, DTCDSCN \cite{liu2020building}, and STANet \cite{chen2020spatial}) and transformer-based methods.
On LEVIR-CD, we notice that DTCDSCN \cite{liu2020building} exhibits 78.05\% IoU. Whereas among recent transformer-based approaches including BIT \cite{chen2021remote_bit}, ChangeFormer \cite{bandara2022transformer}, and TransUNetCD \cite{li2022transunetcd} provide  IoU scores of 80.68\%, 82.48\%, and 83.67\%, respectively. Our method achieves state-of-the-art performance against existing methods and achieves a score of 84.94\%. 

In case of CDD-CD, CNN-based DASNet \cite{chen2020dasnet} obtains an IoU score of 86.39\% while transformer-based TransUNetCD obtains a 94.40\% IoU score. Compared to these, our method also achieves superior performance on the CDD-CD dataset compared to the existing CNN-based and transformer-based approaches and obtains a promising IoU score of 95.41\%.

\subsection{Qualitative Comparison:}
We also present the qualitative comparison of our method over five methods in Fig. \ref{fig_levir_cdd_vis}. We notice that our method demonstrates better capability to capture subtle and large structural information of the change regions compared to existing methods. This validates that our method can better capture both local and global representations for the CD task.

\subsection{Ablation Study:}
Table \ref{tbl:ablation_levir} presents the ablation study of our method. Our baseline, which utilizes ResNet50 \cite{kaiming2016_resnet} features, provides an IoU of 83.08\%.   To better handle the scale variations, we employ multi-scale features from the backbone network, which results in a better IoU score (row 2).
We notice that introducing the convolutional change encoding (CCE) into row 2 further increases the performance (row 3). We also observe a similar pattern (row 4) while introducing attentional change encoding (ACE) into row 2.
Our final method utilizing the proposed ACE and CCE (row 5) further increases the performance and achieves an IoU score of 84.94\% validating the effectiveness of the proposed framework.

\section{Conclusion}
In this work, we present an effective Siamese-based framework (ChangeBind) to better capture the semantic changes between the bi-temporal RS images. To do so, we propose a change encoder that performs convolutional operations to capture the subtle change regions and self-attention operations to exploit the global representations for better CD. Furthermore, we utilize multi-scale features within the change encoder which further improves the performance. Our extensive experimental study over two challenging CD datasets demonstrates that our method has better capability to capture semantic changes and achieve state-of-the-art performance.

\vfill
\pagebreak

\bibliographystyle{IEEEbib}
\bibliography{strings,arxiv}

\begin{thebibliography}{10}

\bibitem{Nuissl2021_urbanisation_and_land_use}
Henning Nuissl and Stefan Siedentop,
\newblock {\em Urbanisation and Land Use Change}, pp. 75--99,
\newblock Springer International Publishing, Cham, 2021.

\bibitem{pang2022cd}
Lei Pang, Jinjin Sun, Yancheng Chi, Yongwen Yang, Fengli Zhang, and Lu~Zhang,
\newblock ``Cd-transunet: A hybrid transformer network for the change detection of urban buildings using l-band sar images,''
\newblock {\em Sustainability}, vol. 14, no. 16, pp. 9847, 2022.

\bibitem{shi2020change}
Wenzhong Shi, Min Zhang, Rui Zhang, Shanxiong Chen, and Zhao Zhan,
\newblock ``Change detection based on artificial intelligence: State-of-the-art and challenges,''
\newblock {\em Remote Sensing}, vol. 12, no. 10, pp. 1688, 2020.

\bibitem{elgcnet2024}
Mubashir Noman, Mustansar Fiaz, Hisham Cholakkal, Salman Khan, and Fahad~Shahbaz Khan,
\newblock ``Elgc-net: Efficient local–global context aggregation for remote sensing change detection,''
\newblock {\em IEEE Transactions on Geoscience and Remote Sensing}, vol. 62, pp. 1--11, 2024.

\bibitem{kucharczyk2021remote}
Maja Kucharczyk and Chris~H Hugenholtz,
\newblock ``Remote sensing of natural hazard-related disasters with small drones: Global trends, biases, and research opportunities,''
\newblock {\em Remote Sensing of Environment}, vol. 264, pp. 112577, 2021.

\bibitem{noman2023remote}
Mubashir Noman, Mustansar Fiaz, Hisham Cholakkal, Sanath Narayan, Rao~Muhammad Anwer, Salman Khan, and Fahad~Shahbaz Khan,
\newblock ``Remote sensing change detection with transformers trained from scratch,''
\newblock {\em IEEE Transactions on Geoscience and Remote Sensing}, 2024.

\bibitem{fonseca2021pattern}
Leila~MG Fonseca, Thales~S K{\"o}rting, Hugo do~N Bendini, Cesare~D Girolamo-Neto, Alana~K Neves, Anderson~R Soares, Evandro~C Taquary, and Raian~V Maretto,
\newblock ``Pattern recognition and remote sensing techniques applied to land use and land cover mapping in the brazilian savannah,''
\newblock {\em Pattern recognition letters}, vol. 148, pp. 54--60, 2021.

\bibitem{wen2021change}
Dawei Wen, Xin Huang, Francesca Bovolo, Jiayi Li, Xinli Ke, Anlu Zhang, and Jon~Atli Benediktsson,
\newblock ``Change detection from very-high-spatial-resolution optical remote sensing images: Methods, applications, and future directions,''
\newblock {\em IEEE Geoscience and Remote Sensing Magazine}, vol. 9, no. 4, pp. 68--101, 2021.

\bibitem{yin2021integrating}
Jiadi Yin, Jinwei Dong, Nicholas~AS Hamm, Zhichao Li, Jianghao Wang, Hanfa Xing, and Ping Fu,
\newblock ``Integrating remote sensing and geospatial big data for urban land use mapping: A review,''
\newblock {\em International Journal of Applied Earth Observation and Geoinformation}, vol. 103, pp. 102514, 2021.

\bibitem{chughtai2021review}
Ali~Hassan Chughtai, Habibullah Abbasi, and Ismail~Rakip Karas,
\newblock ``A review on change detection method and accuracy assessment for land use land cover,''
\newblock {\em Remote Sensing Applications: Society and Environment}, vol. 22, pp. 100482, 2021.

\bibitem{zheng2021building}
Zhuo Zheng, Yanfei Zhong, Junjue Wang, Ailong Ma, and Liangpei Zhang,
\newblock ``Building damage assessment for rapid disaster response with a deep object-based semantic change detection framework: From natural disasters to man-made disasters,''
\newblock {\em Remote Sensing of Environment}, vol. 265, pp. 112636, 2021.

\bibitem{slingsby2020near}
Jasper~A Slingsby, Glenn~R Moncrieff, and Adam~M Wilson,
\newblock ``Near-real time forecasting and change detection for an open ecosystem with complex natural dynamics,''
\newblock {\em ISPRS Journal of Photogrammetry and Remote Sensing}, vol. 166, pp. 15--25, 2020.

\bibitem{migas2021assessment}
Robert Migas-Mazur, Marlena Kycko, Tomasz Zwijacz-Kozica, and Bogdan Zagajewski,
\newblock ``Assessment of sentinel-2 images, support vector machines and change detection algorithms for bark beetle outbreaks mapping in the tatra mountains,''
\newblock {\em Remote sensing}, vol. 13, no. 16, pp. 3314, 2021.

\bibitem{zhu2014continuous}
Zhe Zhu and Curtis~E Woodcock,
\newblock ``Continuous change detection and classification of land cover using all available landsat data,''
\newblock {\em Remote sensing of Environment}, vol. 144, pp. 152--171, 2014.

\bibitem{ghosh2011fuzzy}
Ashish Ghosh, Niladri~Shekhar Mishra, and Susmita Ghosh,
\newblock ``Fuzzy clustering algorithms for unsupervised change detection in remote sensing images,''
\newblock {\em Information Sciences}, vol. 181, no. 4, pp. 699--715, 2011.

\bibitem{hussain2013change}
Masroor Hussain, Dongmei Chen, Angela Cheng, Hui Wei, and David Stanley,
\newblock ``Change detection from remotely sensed images: From pixel-based to object-based approaches,''
\newblock {\em ISPRS Journal of photogrammetry and remote sensing}, vol. 80, pp. 91--106, 2013.

\bibitem{tewkesbury2015critical}
Andrew~P Tewkesbury, Alexis~J Comber, Nicholas~J Tate, Alistair Lamb, and Peter~F Fisher,
\newblock ``A critical synthesis of remotely sensed optical image change detection techniques,''
\newblock {\em Remote Sensing of Environment}, vol. 160, pp. 1--14, 2015.

\bibitem{chen2020dasnet}
Jie Chen, Ziyang Yuan, Jian Peng, Li~Chen, Haozhe Huang, Jiawei Zhu, Yu~Liu, and Haifeng Li,
\newblock ``Dasnet: Dual attentive fully convolutional siamese networks for change detection in high-resolution satellite images,''
\newblock {\em IEEE Journal of Selected Topics in Applied Earth Observations and Remote Sensing}, vol. 14, pp. 1194--1206, 2020.

\bibitem{fang2021snunet}
Sheng Fang, Kaiyu Li, Jinyuan Shao, and Zhe Li,
\newblock ``Snunet-cd: A densely connected siamese network for change detection of vhr images,''
\newblock {\em IEEE Geoscience and Remote Sensing Letters}, vol. 19, pp. 1--5, 2021.

\bibitem{chen2021remote_bit}
Hao Chen, Zipeng Qi, and Zhenwei Shi,
\newblock ``Remote sensing image change detection with transformers,''
\newblock {\em IEEE Transactions on Geoscience and Remote Sensing}, vol. 60, pp. 1--14, 2021.

\bibitem{yan2022fully_ftn}
Tianyu Yan, Zifu Wan, and Pingping Zhang,
\newblock ``Fully transformer network for change detection of remote sensing images,''
\newblock {\em arXiv preprint arXiv:2210.00757}, 2022.

\bibitem{bandara2022transformer}
Wele Gedara~Chaminda Bandara and Vishal~M Patel,
\newblock ``A transformer-based siamese network for change detection,''
\newblock {\em arXiv preprint arXiv:2201.01293}, 2022.

\bibitem{li2022transunetcd}
Qingyang Li, Ruofei Zhong, Xin Du, and Yu~Du,
\newblock ``Transunetcd: A hybrid transformer network for change detection in optical remote-sensing images,''
\newblock {\em IEEE Transactions on Geoscience and Remote Sensing}, vol. 60, pp. 1--19, 2022.

\bibitem{ke2022hybrid_transcd}
Qingtian Ke and Peng Zhang,
\newblock ``Hybrid-transcd: A hybrid transformer remote sensing image change detection network via token aggregation,''
\newblock {\em ISPRS International Journal of Geo-Information}, vol. 11, no. 4, pp. 263, 2022.

\bibitem{kaiming2016_resnet}
Kaiming He, Xiangyu Zhang, Shaoqing Ren, and Jian Sun,
\newblock ``Deep residual learning for image recognition,''
\newblock in {\em 2016 IEEE Conference on Computer Vision and Pattern Recognition (CVPR)}, 2016, pp. 770--778.

\bibitem{daudt2018fully}
Rodrigo~Caye Daudt, Bertr Le~Saux, and Alexandre Boulch,
\newblock ``Fully convolutional siamese networks for change detection,''
\newblock in {\em 2018 25th IEEE International Conference on Image Processing (ICIP)}. IEEE, 2018, pp. 4063--4067.

\bibitem{liu2020building}
Yi~Liu, Chao Pang, Zongqian Zhan, Xiaomeng Zhang, and Xue Yang,
\newblock ``Building change detection for remote sensing images using a dual-task constrained deep siamese convolutional network model,''
\newblock {\em IEEE Geoscience and Remote Sensing Letters}, vol. 18, no. 5, pp. 811--815, 2020.

\bibitem{chen2020spatial}
Hao Chen and Zhenwei Shi,
\newblock ``A spatial-temporal attention-based method and a new dataset for remote sensing image change detection,''
\newblock {\em Remote Sensing}, vol. 12, no. 10, pp. 1662, 2020.

\bibitem{lin2017feature}
Tsung-Yi Lin, Piotr Doll{\'a}r, Ross Girshick, Kaiming He, Bharath Hariharan, and Serge Belongie,
\newblock ``Feature pyramid networks for object detection,''
\newblock in {\em Proceedings of the IEEE conference on computer vision and pattern recognition}, 2017, pp. 2117--2125.

\bibitem{Lebedev2018CHANGEDI}
Maxim Lebedev, Yu.~V. Vizilter, Oleg Vygolov, Vladimir~A. Knyaz, and A.~Yu. Rubis,
\newblock ``Change detection in remote sensing images using conditional adversarial networks,''
\newblock {\em The International Archives of the Photogrammetry, Remote Sensing and Spatial Information Sciences}, 2018.

\end{thebibliography}

\end{document}